\documentclass{article}
\usepackage{ijcai18}

\usepackage{times}
\usepackage{xcolor}
\usepackage{soul}
\usepackage[utf8]{inputenc}
\usepackage[small]{caption}

\usepackage{amsmath}
\usepackage{amsfonts}
\usepackage{epsfig}
\usepackage{graphicx}

\title{Visual Interpretability for Deep Learning: a Survey}

\author{Quanshi Zhang and Song-Chun Zhu\\University of California, Los Angeles}

\begin{document}

\maketitle

\begin{abstract}
This paper reviews recent studies in understanding neural-network representations and learning neural networks with interpretable/disentangled middle-layer representations. Although deep neural networks have exhibited superior performance in various tasks, the interpretability is always the Achilles' heel of deep neural networks. At present, deep neural networks obtain high discrimination power at the cost of low interpretability of their black-box representations. We believe that high model interpretability may help people to break several bottlenecks of deep learning, \emph{e.g.} learning from very few annotations, learning via human-computer communications at the semantic level, and semantically debugging network representations. We focus on convolutional neural networks (CNNs), and we revisit the visualization of CNN representations, methods of diagnosing representations of pre-trained CNNs, approaches for disentangling pre-trained CNN representations, learning of CNNs with disentangled representations, and middle-to-end learning based on model interpretability. Finally, we discuss prospective trends in explainable artificial intelligence.
\end{abstract}

\section{Introduction}

Convolutional neural networks (CNNs)~\cite{CNN,CNNImageNet,ResNet,denseNet} have achieved superior performance in many visual tasks, such as object classification and detection. However, the end-to-end learning strategy makes CNN representations a black box. Except for the final network output, it is difficult for people to understand the logic of CNN predictions hidden inside the network. In recent years, a growing number of researchers have realized that high model interpretability is of significant value in both theory and practice and have developed models with interpretable knowledge representations.

In this paper, we conduct a survey of current studies in understanding neural-network representations and learning neural networks with interpretable/disentangled representations. We can roughly define the scope of the review into the following six research directions.
\begin{itemize}
\item Visualization of CNN representations in intermediate network layers. These methods mainly synthesize the image that maximizes the score of a given unit in a pre-trained CNN or invert feature maps of a conv-layer back to the input image. Please see Section~\ref{sec:visualization} for detailed discussion.

\item Diagnosis of CNN representations. Related studies may either diagnose a CNN's feature space for different object categories or discover potential representation flaws in conv-layers. Please see Section~\ref{sec:diagnosis} for details.

\item Disentanglement of ``the mixture of patterns'' encoded in each filter of CNNs. These studies mainly disentangle complex representations in conv-layers and transform network representations into interpretable graphs. Please see Section~\ref{sec:explanatory} for details.

\item Building explainable models. We discuss interpretable CNNs~\cite{InterpretableCNN}, capsule networks~\cite{capsule}, interpretable R-CNNs~\cite{InterRCNN}, and the InfoGAN~\cite{InfoGAN} in Section~\ref{sec:interpretableNets}.

\item Semantic-level middle-to-end learning via human-computer interaction. A clear semantic disentanglement of CNN representations may further enable ``middle-to-end'' learning of neural networks with weak supervision. Section~\ref{sec:middletoend} introduces methods to learn new models via human-computer interactions~\cite{holdingHands} and active question-answering with very limited human supervision~\cite{DeepQA}.
\end{itemize}
Among all the above, the visualization of CNN representations is the most direct way to explore network representations. The network visualization also provides a technical foundation for many approaches to diagnosing CNN representations. The disentanglement of feature representations of a pre-trained CNN and the learning of explainable network representations present bigger challenges to state-of-the-art algorithms. Finally, explainable or disentangled network representations are also the starting point for weakly-supervised middle-to-end learning.

\textbf{Values of model interpretability:} The clear semantics in high conv-layers can help people trust a network's prediction. As discussed in \cite{CNNBias}, considering dataset and representation bias, a high accuracy on testing images still cannot ensure that a CNN will encode correct representations. For example, a CNN may use an unreliable context---eye features---to identify the ``lipstick'' attribute of a face image. Therefore, people usually cannot fully trust a network unless a CNN can semantically or visually explain its logic, \emph{e.g.} what patterns are used for prediction.

In addition, the middle-to-end learning or debugging of neural networks based on the explainable or disentangled network representations may also significantly reduce the requirement for human annotation. Furthermore, based on semantic representations of networks, it is possible to merge multiple CNNs into a universal network (\emph{i.e.} a network encoding generic knowledge representations for different tasks) at the semantic level in the future.

In the following sections, we review the above research directions and discuss the potential future of technical developments.

\section{Visualization of CNN representations}
\label{sec:visualization}

Visualization of filters in a CNN is the most direct way of exploring visual patterns hidden inside a neural unit. Different types of visualization methods have been developed for network visualization.

First, gradient-based methods~\cite{CNNVisualization_1,CNNVisualization_2,CNNVisualization_3,deconvnet} are the mainstream of network visualization. These methods mainly compute gradients of the score of a given CNN unit \emph{w.r.t.} the input image. They use the gradients to estimate the image appearance that maximizes the unit score. \cite{olah2017feature} has provided a toolbox of existing techniques to visualize patterns encoded in different conv-layers of a pre-trained CNN.

Second, the up-convolutional net~\cite{FeaVisual} is another typical technique to visualize CNN representations. The up-convolutional net inverts CNN feature maps to images. We can regard up-convolutional nets as a tool that indirectly illustrates the image appearance corresponding to a feature map, although compared to gradient-based methods, up-convolutional nets cannot mathematically ensure that the visualization result exactly reflects actual representations in the CNN. Similarly, \cite{nguyen2017plug} has further introduced an additional prior, which controls the semantic meaning of the synthesized image, to the adversarial generative network. We can use CNN feature maps as the prior for visualization.

In addition, \cite{CNNSemanticDeep} has proposed a method to accurately compute the image-resolution receptive field of neural activations in a feature map. The actual receptive field of neural activation is smaller than the theoretical receptive field computed using the filter size. The accurate estimation of the receptive field helps people to understand the representation of a filter.

\section{Diagnosis of CNN representations}
\label{sec:diagnosis}

Some methods go beyond the visualization of CNNs and diagnose CNN representations to obtain insight understanding of features encoded in a CNN. We roughly divide all relevant research into the following five directions.

Studies in the first direction analyze CNN features from a global view. \cite{CNNAnalysis_1} has explored semantic meanings of each filter. \cite{CNNAnalysis_2} has analyzed the transferability of filter representations in intermediate conv-layers. \cite{CNNAnalysis_3,CNNVisualization_5} have computed feature distributions of different categories/attributes in the feature space of a pre-trained CNN.

The second research direction extracts image regions that directly contribute the network output for a label/attribute to explain CNN representations of the label/attribute. This is similar to the visualization of CNNs. Methods of \cite{visualCNN_grad,visualCNN_grad_2} have been proposed to propagate gradients of feature maps \emph{w.r.t.} the final loss back to the image plane to estimate the image regions. The LIME model proposed in \cite{trust} extracts image regions that are highly sensitive to the network output. Studies of \cite{VisualizationDifference,patternNet,ExplainingArea} have invented methods to visualize areas in the input image that contribute the most to the decision-making process of the CNN. \cite{QiWuVQA,interpretVQA_grad} have tried to interpret the logic for visual question-answering encoded in neural networks. These studies list important objects (or regions of interests) detected from the images and crucial words in questions as the explanation of output answers.

The estimation of vulnerable points in the feature space of a CNN is also a popular direction for diagnosing network representations. Approaches of \cite{pixelAttack,CNNInfluence,CNNAnalysis_1} have been developed to compute adversarial samples for a CNN. \emph{I.e.} these studies aim to estimate the minimum noisy perturbation of the input image that can change the final prediction. In particular, influence functions proposed in \cite{CNNInfluence} can be used to compute adversarial samples. The influence function can also provide plausible ways to create training samples to attack the learning of CNNs, fix the training set, and further debug representations of a CNN.

The fourth research direction is to refine network representations based on the analysis of network feature spaces. Given a CNN pre-trained for object classification, \cite{banditUnknown} has proposed a method to discover knowledge blind spots (unknown patterns) of the CNN in a weakly-supervised manner. This method grouped all sample points in the entire feature space of a CNN into thousands of pseudo-categories. It assumed that a well learned CNN would use the sub-space of each pseudo-category to exclusively represent a subset of a specific object class. In this way, this study randomly showed object samples within each sub-space, and used the sample purity in the sub-space to discover potential representation flaws hidden in a pre-trained CNN. To distill representations of a teacher network to a student network for sentiment analysis, \cite{LogicRuleNetwork} has proposed using logic rules of natural languages (\emph{e.g.} I-ORG cannot follow B-PER) to construct a distillation loss to supervise the knowledge distillation of neural networks, in order to obtain more meaningful network representations.

\begin{figure}[t]
\centering
\includegraphics[width=\linewidth]{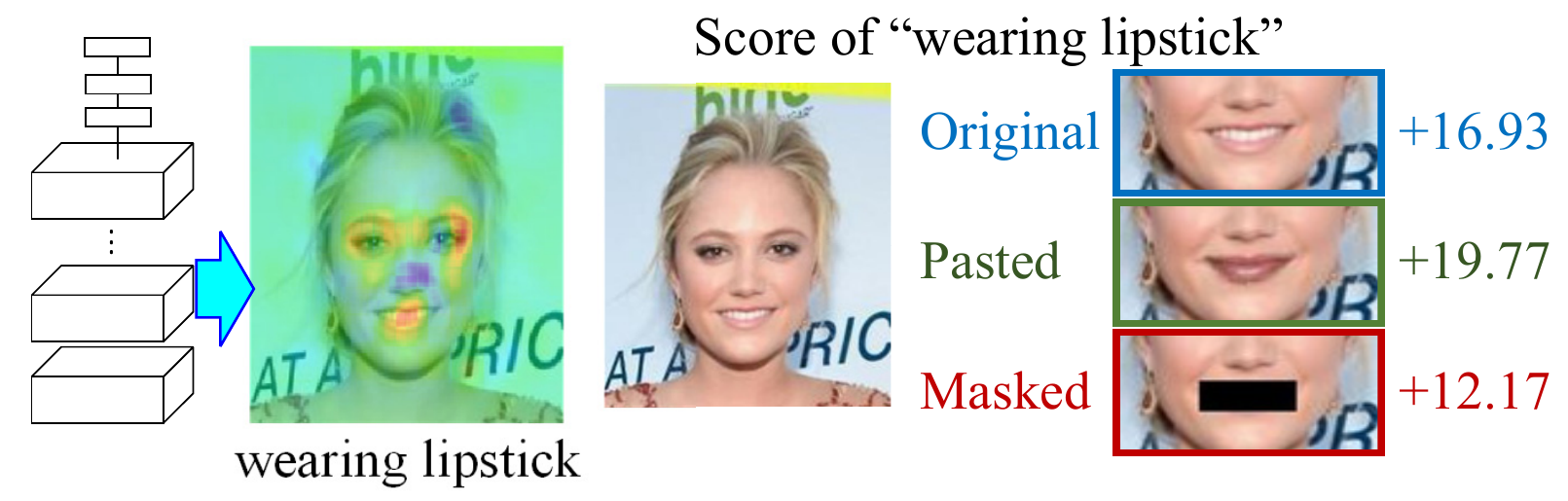}
\caption{Biased representations in a CNN~{\protect\cite{CNNBias}}. Considering potential dataset bias, a high accuracy on testing images cannot always ensure that a CNN learns correct representations. The CNN may use unreliable co-appearing contexts to make predictions. For example, people may manually modify mouth appearances of two faces by masking mouth regions or pasting another mouth, but such modifications do not significantly change prediction scores for the \textit{lipstick} attribute. This figure shows heat maps of inference patterns of the \textit{lipstick} attribute, where patterns with red/blue colors are positive/negative with the attribute score. The CNN mistakenly considers unrelated patterns as contexts to infer the lipstick.}
\label{fig:bias}
\end{figure}

Finally, \cite{CNNBias} has presented a method to discover potential, biased representations of a CNN. Fig.~\ref{fig:bias} shows biased representations of a CNN trained for the estimation of face attributes. When an attribute usually co-appears with specific visual features in training images, then the CNN may use such co-appearing features to represent the attribute. When the used co-appearing features are not semantically related to the target attribute, these features can be considered as biased representations.

Given a pre-trained CNN (\emph{e.g.} a CNN that was trained to estimate face attributes), \cite{CNNBias} required people to annotate some ground-truth relationships between attributes, \emph{e.g.} the \textit{lipstick} attribute is positively related to the \textit{heavy-makeup} attribute, and is not related to the \textit{black hair} attribute. Then, the method mined inference patterns of each attribute output from conv-layers, and used inference patterns to compute actual attribute relationships encoded in the CNN. Conflicts between the ground-truth and the mined attribute relationships indicated biased representations.

\section{Disentangling CNN representations into explanatory graphs \& decision trees}
\label{sec:explanatory}

\subsection{Disentangling CNN representations into explanatory graphs}

Compared to the visualization and diagnosis of network representations in previous sections, disentangling CNN features into human-interpretable graphical representations (namely \textit{explanatory graphs}) provides a more thorough explanation of network representations. \cite{explanatoryGraph,CNNAoG} have proposed disentangling features in conv-layers of a pre-trained CNN and have used a graphical model to represent the semantic hierarchy hidden inside a CNN.

\begin{figure}[t]
\centering
\includegraphics[width=\linewidth]{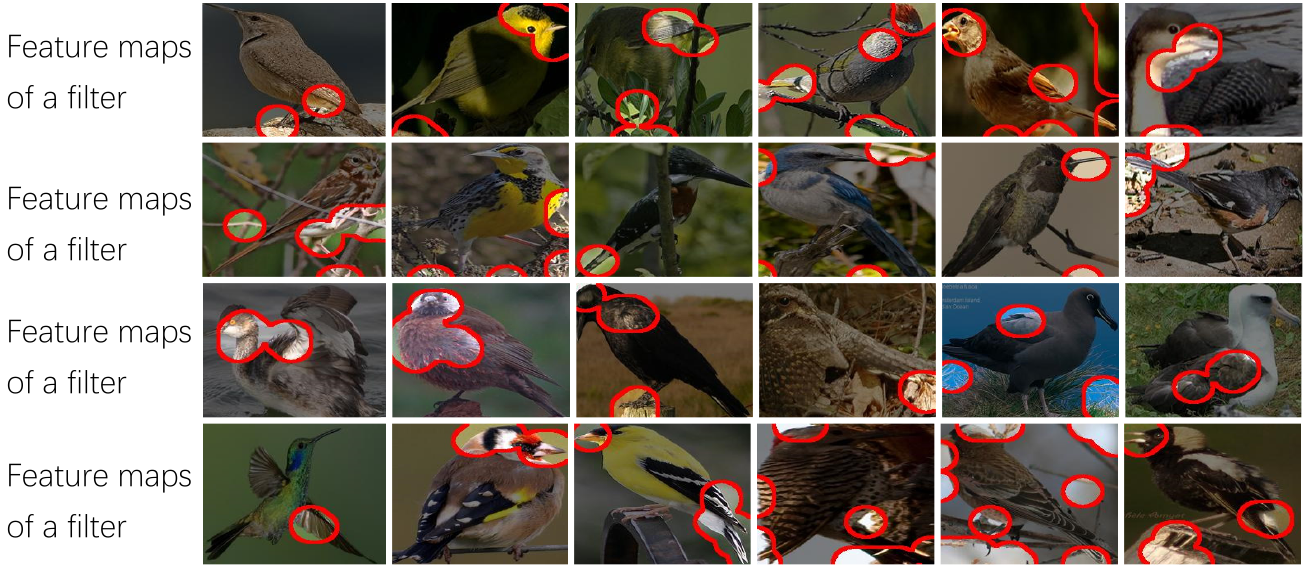}
\caption{Feature maps of a filter obtained using different input images~{\protect\cite{explanatoryGraph}}. To visualize the feature map, the method propagates receptive fields of activated units in the feature map back to the image plane. In each sub-feature, the filter is activated by various part patterns in an image. This makes it difficult to understand the semantic meaning of a filter.}
\label{fig:mix}
\end{figure}

As shown in Fig.~\ref{fig:mix}, each filter in a high conv-layer of a CNN usually represents a mixture of patterns. For example, the filter may be activated by both the head and the tail parts of an object. Thus, to provide a global view of how visual knowledge is organized in a pre-trained CNN, studies of \cite{explanatoryGraph,CNNAoG} aim to answer the following three questions.
\begin{itemize}
\item How many types of visual patterns are memorized by each convolutional filter of the CNN (here, a visual pattern may describe a specific object part or a certain texture)?
\item Which patterns are co-activated to describe an object part?
\item What is the spatial relationship between two co-activated patterns?
\end{itemize}

\begin{figure}[t]
\centering
\includegraphics[width=\linewidth]{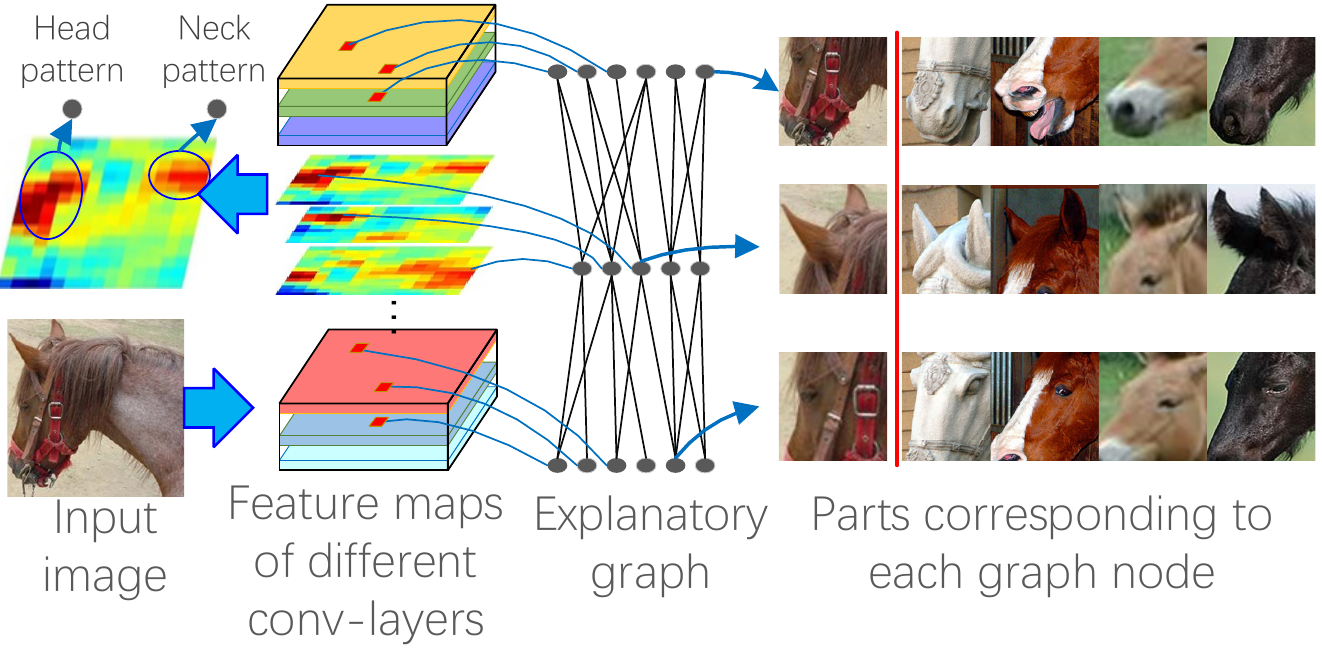}
\caption{Explanatory graph~{\protect\cite{explanatoryGraph}}. An explanatory graph represents the knowledge hierarchy hidden in conv-layers of a CNN. Each filter in a pre-trained CNN may be activated by different object parts. {\protect\cite{explanatoryGraph}} disentangles part patterns from each filter in an unsupervised manner, thereby clarifying the knowledge representation.}
\label{fig:explanatory}
\end{figure}

\begin{figure*}[t]
\centering
\includegraphics[width=0.8\linewidth]{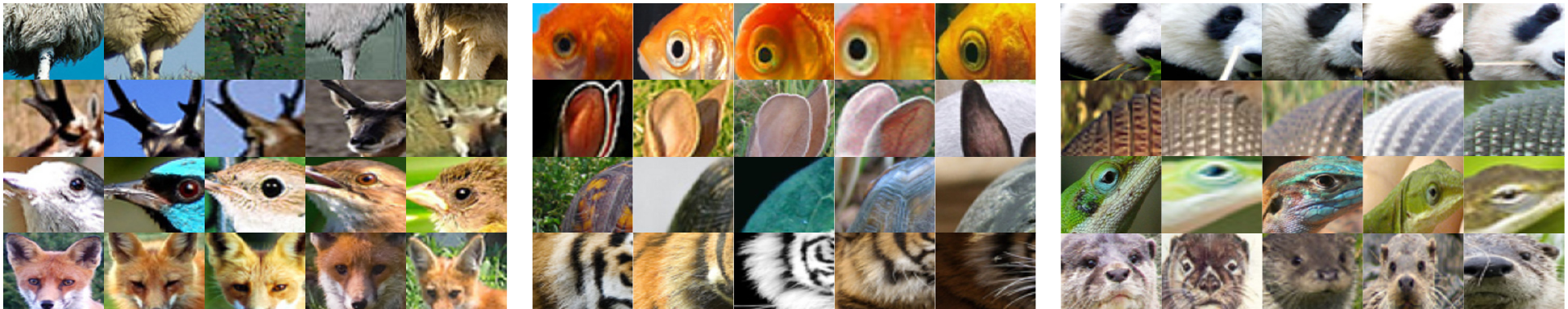}
\caption{Image patches corresponding to different nodes in the explanatory graph~{\protect\cite{explanatoryGraph}}.}
\label{fig:patch}
\end{figure*}

\begin{figure*}[t]
\centering
\includegraphics[width=0.7\linewidth]{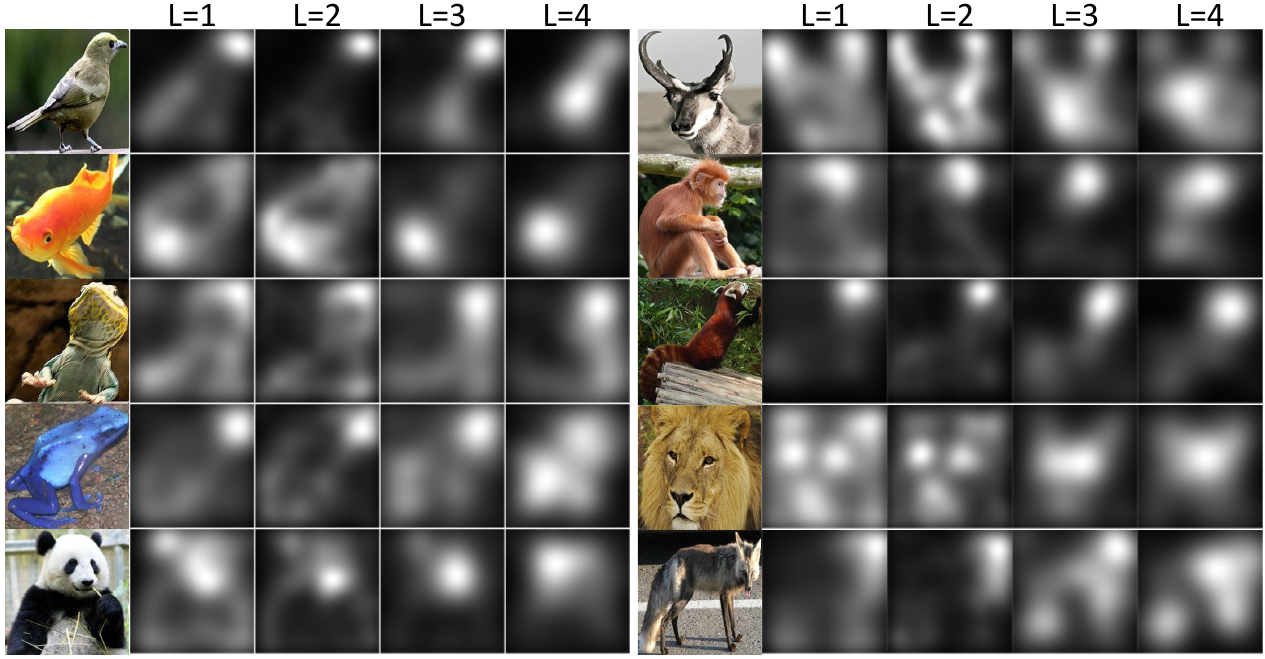}
\caption{Heat maps of patterns~{\protect\cite{explanatoryGraph}}. A heat map visualizes the spatial distribution of the top 50\% patterns in the $L$-th layer of the explanatory graph with the highest inference scores.}
\label{fig:heatmap}
\end{figure*}

\begin{figure*}[t]
\centering
\includegraphics[width=0.9\linewidth]{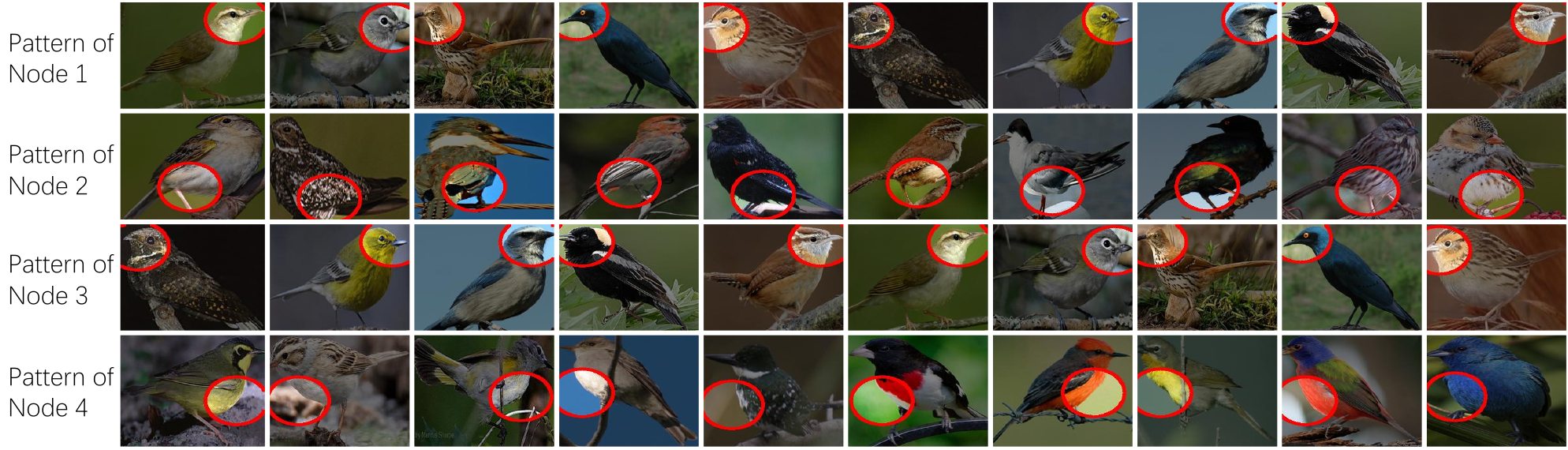}
\caption{Image regions inferred by each node in an explanatory graph~{\protect\cite{explanatoryGraph}}. The method of {\protect\cite{explanatoryGraph}} successfully disentangles object-part patterns from representations of every single filter.}
\label{fig:interpretability}
\end{figure*}

As shown in Fig.~\ref{fig:explanatory}, the explanatory graph explains the knowledge semantic hidden inside the CNN. The explanatory graph disentangles the mixture of part patterns in each filter's feature map of a conv-layer, and uses each graph node to represent a part.
\begin{itemize}
\item The explanatory graph has multiple layers. Each graph layer corresponds to a specific conv-layer of a CNN.
\item Each filter in a conv-layer may represent the appearance of different object parts. The algorithm automatically disentangles the mixture of part patterns encoded in a single filter, and uses a node in the explanatory graph to represent each part pattern.
\item Each node in the explanatory graph consistently represents the same object part through different images. We can use the node to localize the corresponding part on the input image. To some extent, the node is robust to shape deformation and pose variations.
\item Each edge encodes the co-activation relationship and the spatial relationship between two nodes in adjacent layers.
\item We can regard an explanatory graph as a compression of feature maps of conv-layers. A CNN has multiple conv-layers. Each conv-layer may have hundreds of filters, and each filter may produce a feature map with hundreds of neural units. We can use tens of thousands of nodes in the explanatory graph to represent information contained in all tens of millions of neural units in these feature maps, \emph{i.e.} by which part patterns the feature maps are activated, and where the part patterns are localized in input images.
\item Just like a dictionary, each input image can only trigger a small subset of part patterns (nodes) in the explanatory graph. Each node describes a common part pattern with high transferability, which is shared by hundreds or thousands of training images.
\end{itemize}

Fig.~\ref{fig:patch} lists top-ranked image patches corresponding to different nodes in the explanatory graph. Fig.~\ref{fig:heatmap} visualizes the spatial distribution of object parts inferred by the top 50\% nodes in the $L$-th layer of the explanatory graph with the highest inference scores. Fig.~\ref{fig:interpretability} shows object parts inferred by a single node.

\subsubsection{Application: multi-shot part localization}

There are many potential applications based on the explanatory graph. For example, we can regard the explanatory graph as a visual dictionary of a category and transfer graph nodes to other applications, such as multi-shot part localization.

Given very few bounding boxes of an object part, \cite{explanatoryGraph} has proposed retrieving hundreds of nodes that are related to the part annotations from the explanatory graph, and then use the retrieved nodes to localize object parts in previously unseen images. Because each node in the explanatory graph encodes a part pattern shared by numerous training images, the retrieved nodes describe a general appearance of the target part without being over-fitted to the limited annotations of part bounding boxes. Given three annotations for each object part, the explanatory-graph-based method has exhibited superior performance of part localization and has decreased by about 1/3 localization errors \emph{w.r.t.} the second-best baseline.

\subsection{Disentangling CNN representations into decision trees}

\begin{figure}[t]
\centering
\includegraphics[width=0.99\linewidth]{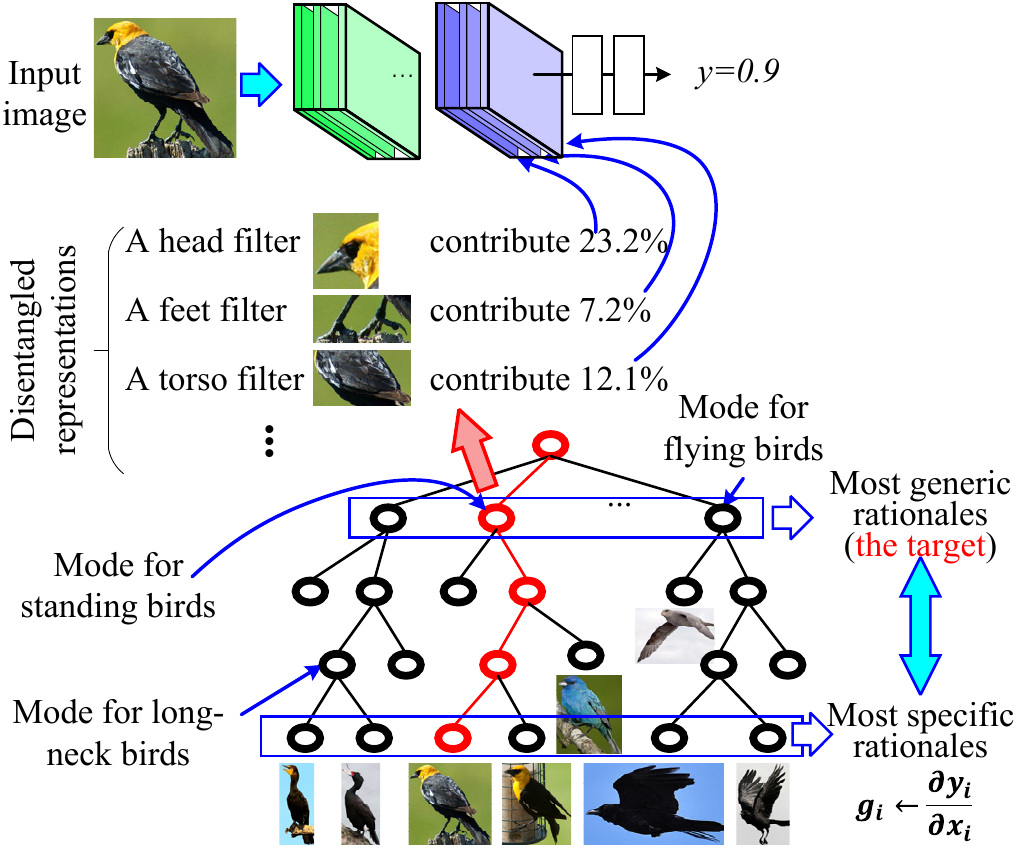}
\caption{Decision tree that explains a CNN prediction at the semantic level~{\protect\cite{explanatoryTree}}. A CNN is learned for object classification with disentangled representations in the top conv-layer, where each filter represents a specific object part. The decision tree encodes various decision modes hidden inside fully-connected layers of the CNN in a coarse-to-fine manner. Given an input image, the decision tree infers a parse tree (red lines) to quantitatively analyze rationales for the CNN prediction, \emph{i.e.} which object parts (or filters) are used for prediction and how much an object part (or filter) contributes to the prediction.}
\label{fig:expTree}
\end{figure}

\cite{explanatoryTree} has further proposed a decision tree to encode decision modes in fully-connected layers. The decision tree is not designed for classification. Instead, the decision tree is used to quantitatively explain the logic for each CNN prediction. \emph{I.e.} given an input image, we use the CNN to make a prediction. The decision tree tells people which filters in a conv-layer are used for the prediction and how much they contribute to the prediction.

As shown in Fig.~\ref{fig:expTree}, the method mines potential decision modes memorized in fully-connected layers. The decision tree organizes these potential decision modes in a coarse-to-fine manner. Furthermore, this study uses the method of \cite{InterpretableCNN} to disentangle representations of filters in the top conv-layers, \emph{i.e.} making each filter represent a specific object part. In this way, people can use the decision tree to explain rationales for each CNN prediction at the semantic level, \emph{i.e.} which object parts are used by the CNN to make the prediction.

\section{Learning neural networks with interpretable/disentangled representations}
\label{sec:interpretableNets}

Almost all methods mentioned in previous sections focus on the understanding of a pre-trained network. In this section, we review studies of learning disentangled representations of neural networks, where representations in middle layers are no longer a black box but have clear semantic meanings. Compared to the understanding of pre-trained networks, learning networks with disentangled representations present more challenges. Up to now, only a few studies have been published in this direction.

\begin{figure}[t]
\centering
\includegraphics[width=\linewidth]{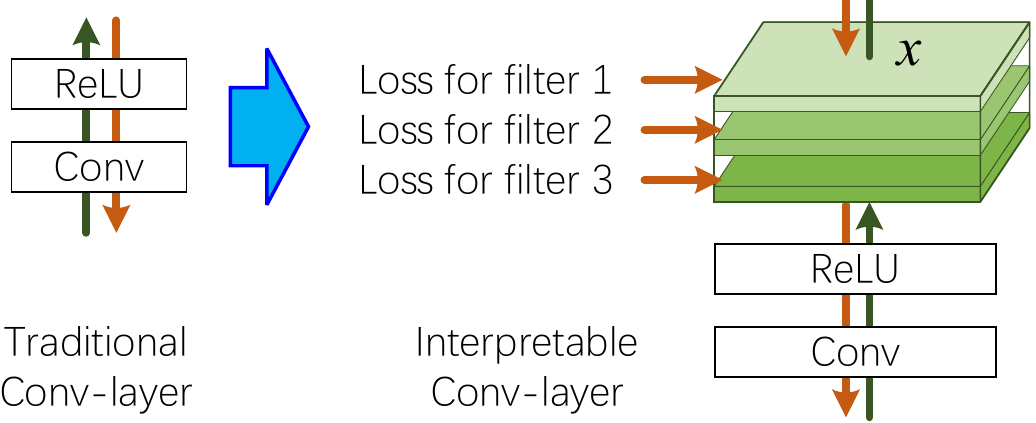}
\caption{Structures of an ordinary conv-layer and an interpretable conv-layer~{\protect\cite{InterpretableCNN}}. Green and red lines indicate the forward and backward propagations, respectively.}
\label{fig:net}
\includegraphics[width=\linewidth]{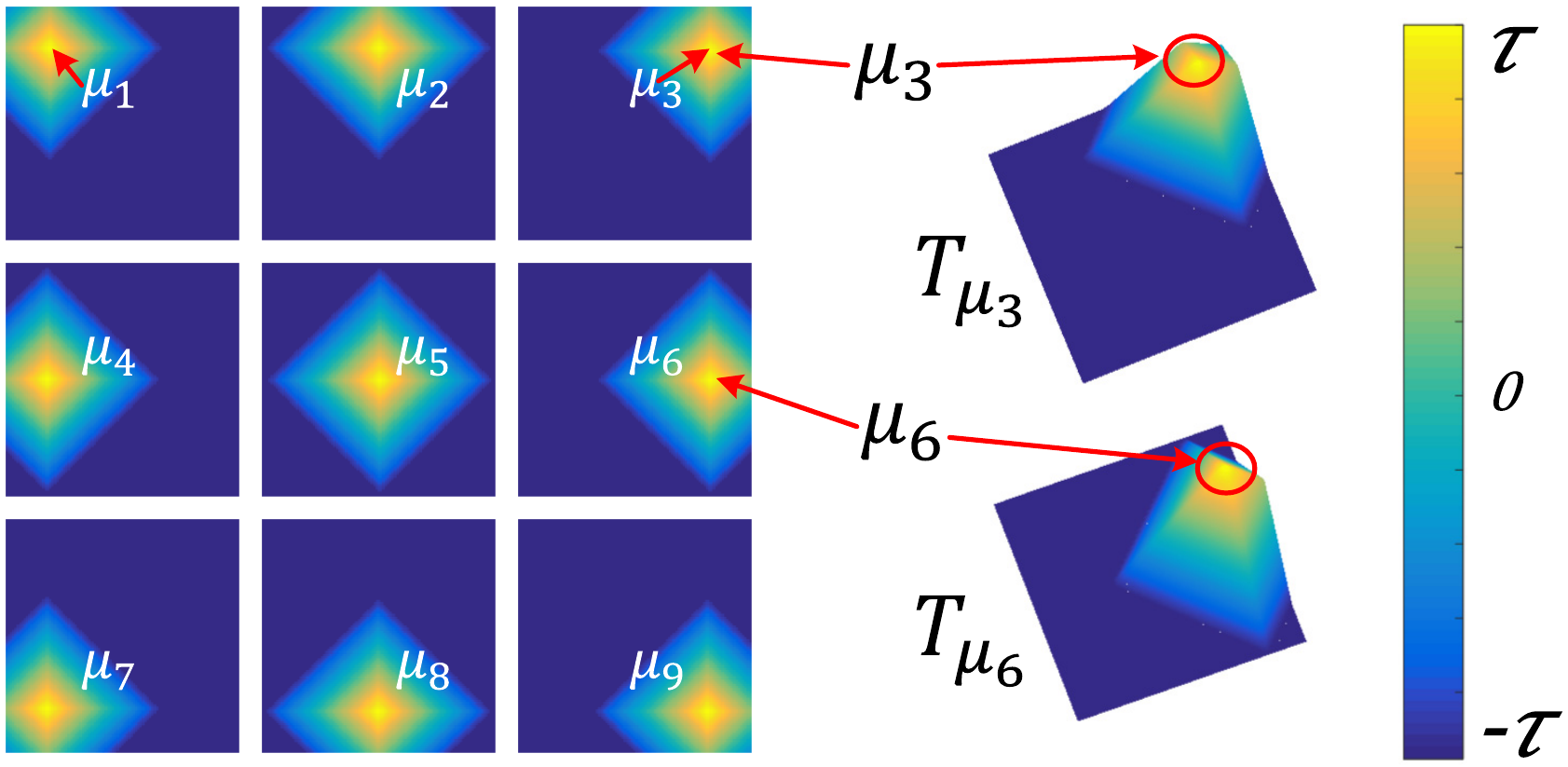}
\caption{Templates~{\protect\cite{InterpretableCNN}}. Each template $T_{\mu_{i}}$ matches to a feature map when the target part mainly triggers the $i$-th unit in the feature map.}
\label{fig:template}
\end{figure}

\begin{figure*}[t]
\centering
\includegraphics[width=\linewidth]{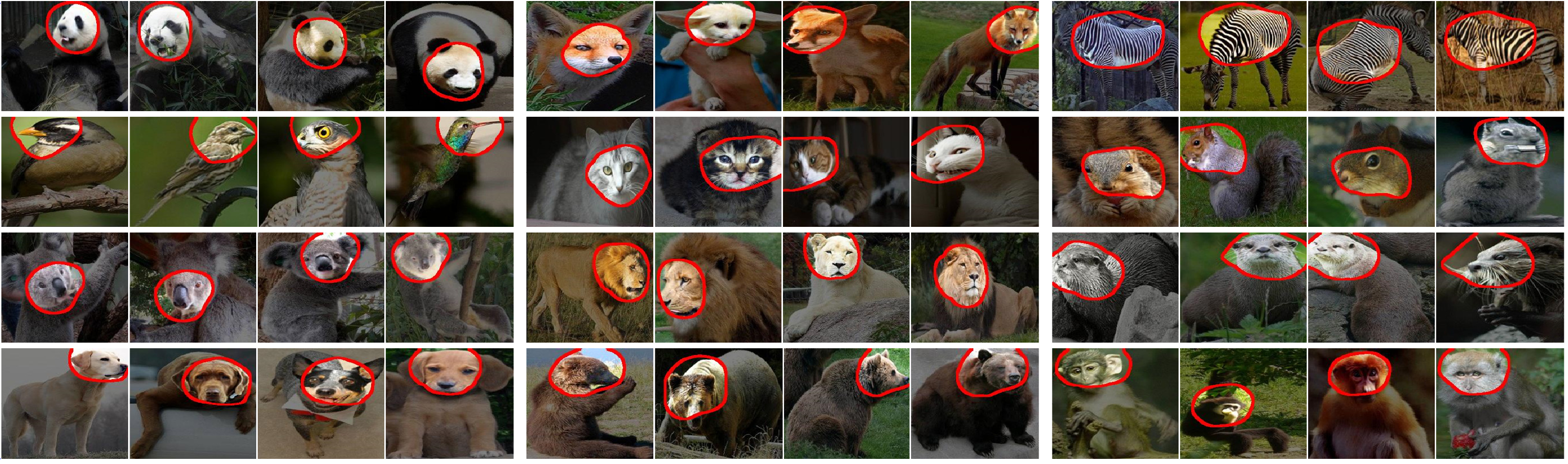}
\caption{Visualization of interpretable filters in the top conv-layer~{\protect\cite{InterpretableCNN}}. We used {\protect\cite{CNNSemanticDeep}} to estimate the image-resolution receptive field of activations in a feature map to visualize a filter's semantics. An interpretable CNN usually encodes head patterns of animals in its top conv-layer for classification.}
\label{fig:interpretablePart}
\end{figure*}

\subsection{Interpretable convolutional neural networks}

As shown in Fig.~\ref{fig:net}, \cite{InterpretableCNN} has developed a method to modify an ordinary CNN to obtain disentangled representations in high conv-layers by adding a loss to each filter in the conv-layers. The loss is used to regularize the feature map towards the representation of a specific object part.

Note that people do not need to annotate any object parts or textures to supervise the learning of interpretable CNNs. Instead, the loss automatically assigns an object part to each filter during the end-to-end learning process. As shown in Fig.~\ref{fig:template}, this method designs some templates. Each template $T_{\mu_{i}}$ is a matrix with the same size of feature map. $T_{\mu_{i}}$ describes the ideal distribution of activations for the feature map when the target part mainly triggers the $i$-th unit in the feature map.

Given the joint probability of fitting a feature map to a template, the loss of a filter is formulated as the mutual information between the feature map and the templates. This loss encourages a low entropy of inter-category activations. \emph{I.e.} each filter in the conv-layer is assigned to a certain category. If the input image belongs to the target category, then the loss expects the filter's feature map to match a template well; otherwise, the filter needs to remain inactivated. In addition, the loss also encourages a low entropy of spatial distributions of neural activations. \emph{I.e.} when the input image belongs the target category, the feature map is supposed to exclusively fit a single template. In other words, the filter needs to activate a single location on the feature map.

This study assumes that if a filter repetitively activates various feature-map regions, then this filter is more likely to describe low-level textures (\emph{e.g.} colors and edges), instead of high-level parts. For example, the left eye and the right eye may be represented by different filters, because contexts of the two eyes are symmetric, but not the same.

Fig.\ref{fig:interpretablePart} shows feature maps produced by different filters of an interpretable CNN. Each filter consistently represents the same object part through various images.

\begin{figure}[t]
\centering
\includegraphics[width=\linewidth]{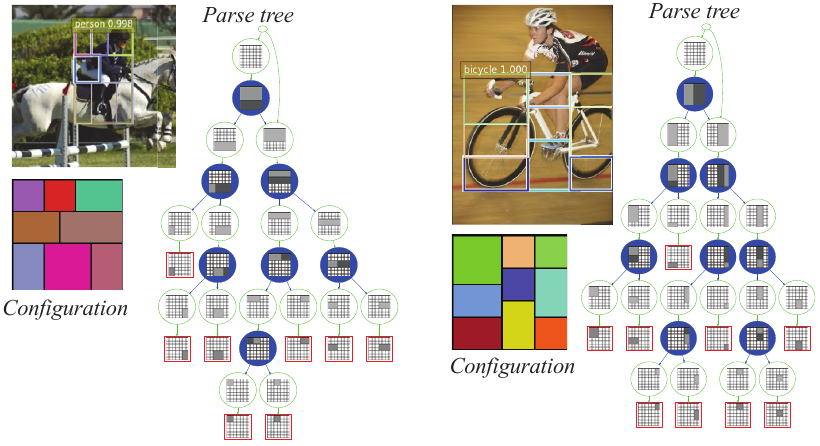}
\caption{Detection examples of the proposed method~{\protect\cite{InterRCNN}}. In addition to predicted bounding boxes, the method also outputs the latent parse tree and part configurations as the qualitatively extractive rationale in detection. The parse trees are inferred on-the-fly in the space of latent structures, which follow a top-down compositional grammar of an AOG.}
\label{fig:AOGParsing}
\end{figure}

\subsection{Interpretable R-CNN}

\cite{InterRCNN} has proposed the learning of qualitatively interpretable models for object detection based on the R-CNN. The objective is to unfold latent configurations of object parts automatically during the object-detection process. This method is learned without using any part annotations for supervision. \cite{InterRCNN} uses a top-down hierarchical and compositional grammar, namely an And-Or graph (AOG), to model latent configurations of object parts. This method uses an AOG-based parsing operator to substitute for the RoI-Pooling operator used in the R-CNN. The AOG-based parsing harnesses explainable compositional structures of objects and maintains the discrimination power of a R-CNN. This idea is related to the disentanglement of the local, bottom-up, and top-down information components for prediction~\cite{wu2007compositional,yang2009evaluating,wu2011numerical}.

During the detection process, a bounding box is interpreted as the best parse tree derived from the AOG on-the-fly. During the learning process, a folding-unfolding method is used to train the AOG and R-CNN in an end-to-end manner.

Fig.~\ref{fig:AOGParsing} illustrates an example of object detection. The proposed method detects object bounding boxes. The method also determines the latent parse tree and part configurations of objects as the qualitatively extractive rationale in detection.

\subsection{Capsule networks}

\cite{capsule} has designed novel neural units, namely capsules, in order to substitute for traditional neural units to construct a capsule network. Each capsule outputs an activity vector instead of a scalar. The length of the activity vector represents the activation strength of the capsule, and the orientation of the activity vector encodes instantiation parameters. Active capsules in the lower layer send messages to capsules in the adjacent higher layer. This method uses an iterative routing-by-agreement mechanism to assign higher weights with the low-layer capsules whose outputs better fit the instantiation parameters of the high-layer capsule.

Experiments showed that when people trained capsule networks using the MNIST dataset~\cite{MNIST}, a capsule encoded a specific semantic concept. Different dimensions of the activity vector of a capsule controlled different features, including 1) scale and thickness, 2) localized part, 3) stroke thickness, 3) localized skew, and 4) width and translation.

\subsection{Information maximizing generative adversarial nets}

The information maximizing generative adversarial net~\cite{InfoGAN}, namely InfoGAN, is an extension of the generative adversarial network. The InfoGAN maximizes the mutual information between certain dimensions of the latent representation and the image observation. The InfoGAN separates input variables of the generator into two types, \emph{i.e.} the incompressible noise $z$ and the latent code $c$. This study aims to learn the latent code $c$ to encode certain semantic concepts in an unsupervised manner.

The InfoGAN has been trained using the MNIST dataset~\cite{MNIST}, the CelebA dataset~\cite{CelebA}, the SVHN dataset~\cite{SVHN}, the 3D face dataset~\cite{Dataset3DFace}, and the 3D chair dataset~\cite{Dataset3DChair}. Experiments have shown that the latent code has successfully encoded the digit type, the rotation, and the width of digits in the MNIST dataset, the lighting condition and the plate context in the SVHN dataset, the azimuth, the existence of glasses, the hairstyle, and the emotion in the CelebA dataset, and the width and 3D rotation in the 3D face and chair datasets.

\section{Evaluation metrics for network interpretability}

Evaluation metrics for model interpretability are crucial for the development of explainable models. This is because unlike traditional well-defined visual applications (\emph{e.g.} object detection and segmentation), network interpretability is more difficult to define and evaluate. The evaluation metric of network interpretability can help people define the concept of network interpretability and guide the development of learning interpretable network representations. Up to now, only very few studies have discussed the evaluation of network interpretability. Proposing a promising evaluation metric is still a big challenge to state-of-the-art algorithms. In this section, we simply introduce two latest evaluation metrics for the interpretability of CNN filters, \emph{i.e.} the filter interpretability proposed by \cite{Interpretability} and the location instability proposed by \cite{explanatoryGraph}.

\subsection{Filter interpretability}

\cite{Interpretability} has defined six types of semantics for CNN filters, \emph{i.e.} \textit{objects}, \textit{parts}, \textit{scenes}, \textit{textures}, \textit{materials}, and \textit{colors}. The evaluation of filter interpretability requires people to annotate these six types of semantics on testing images at the pixel level. The evaluation metric measures the fitness between the image-resolution receptive field of a filter's neural activations\footnote{The method propagates the receptive field of each activated unit in a filter's feature map back to the image plane as the image-resolution receptive field of a filter.} and the pixel-level semantic annotations on the image. For example, if the receptive field of a filter's neural activations usually highly overlaps with ground-truth image regions of a specific semantic concept through different images, then we can consider that the filter represents this semantic concept.

For each filter $f$, this method computes its feature maps {\small${\bf X}=\{x=f(I)|I\in{\bf I}\}$} on different testing images. Then, the distribution of activation scores in all positions of all feature maps is computed. \cite{Interpretability} set an activation threshold {\small$T_{f}$} such that {\small$p(x_{ij}>T_{f})=0.005$}, to select top activations from all spatial locations {\small$[i,j]$} of all feature maps {\small$x\in{\bf X}$} as valid map regions corresponding to $f$'s semantics. Then, the method scales up low-resolution valid map regions to the image resolution, thereby obtaining the receptive field of valid activations on each image. We use {\small$S_{f}^{I}$} to denote the receptive field of $f$'s valid activations \emph{w.r.t.} the image $I$.

The compatibility between a filter $f$ and a specific semantic concept is reported as an intersection-over-union score {\small$IoU_{f,k}^{I}\!=\!\frac{\Vert S_{f}^{I}\cap S_{k}^{I}\Vert}{\Vert S_{f}^{I}\cup S_{k}^{I}\Vert}$}, where {\small$S_{k}^{I}$} denotes the ground-truth mask of the $k$-th semantic concept on the image $I$. Given an image $I$, filter $f$ is associated with the $k$-th concept if {\small$IoU_{f,k}^{I}>0.04$}. The probability of the $k$-th concept being associated with the filter $f$ is given as {\small$P_{f,k}={\textrm{mean}}_{I:\textrm{with k-th concept}}{\bf1}(IoU_{f,k}^{I}>0.04)$}. Thus, we can use $P_{f,k}$ to evaluate the filter interpretability of $f$.

\subsection{Location instability}

\begin{figure}[t]
\centering
\includegraphics[width=\linewidth]{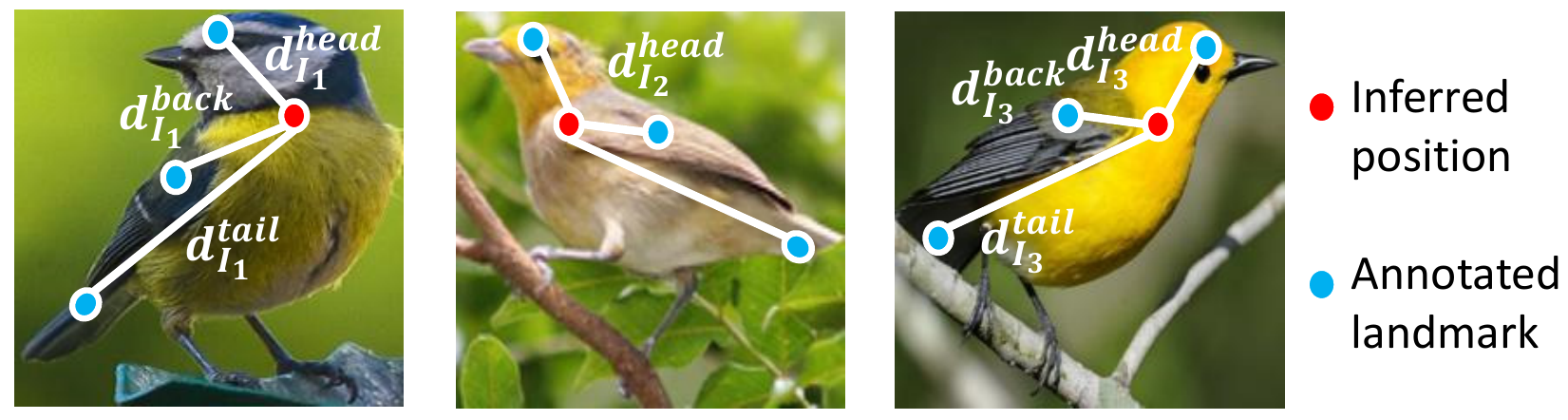}
\caption{Notation for the computation of a filter's location instability~{\protect\cite{explanatoryGraph}}.}
\label{fig:instability}
\end{figure}

Another evaluation metric is location instability. This metric is proposed by \cite{explanatoryGraph} to evaluate the fitness between a CNN filter and the representation of an object part. Given an input image $I$, the CNN computes a feature map $x\in\mathbb{R}^{N\times N}$ of filter $f$. We can regard the unit $x_{i,j}$ ($1\leq i,j\leq N$) with the highest activation as the location inference of $f$, where $N\times N$ is referred to as the size of the feature map. We use $\hat{\bf p}$ to denote the image position that corresponds to the inferred feature map location $(i,j)$, \emph{i.e.} the center of the unit $x_{i,j}$'s receptive field when we backward propagated the receptive field to the image plane. The evaluation assumes that if $f$ consistently represented the same object part (the object part may not have an explicit name according to people's cognition) through different objects, then distances between the image position $\hat{\bf p}$ and some object landmarks should not change much among different objects. For example, if filter $f$ represents the shoulder, then the distance between the shoulder and the head should remain stable through different objects.

Therefore, people can compute the deviation of the distance between the inferred position $\hat{\bf p}$ and a specific ground-truth landmark among different images. The average deviation \emph{w.r.t.} various landmarks can be used to evaluate the location instability of $f$. As shown in Fig.~\ref{fig:instability}, let {\small$d_{I}({\bf p}_{k},\hat{\bf p})=\frac{\Vert {\bf p}_{k}-\hat{\bf p}\Vert}{\sqrt{w^2+h^2}}$} denote the normalized distance between the inferred part and the $k$-th landmark {\small${\bf p}_{k}$} on image $I$. {\small$\sqrt{w^2+h^2}$} denotes the diagonal length of the input image. Thus, {\small$D_{f,k}=\sqrt{{\textrm{var}}_{I}[d_{I}({\bf p}_{k},\hat{\bf p})]}$} is reported as the relative location deviation of filter $f$ \emph{w.r.t.} the $k$-th landmark, where {\small${\textrm{var}}_{I}[d_{I}({\bf p}_{k},\hat{\bf p})]$} is referred to as the variation of the distance {\small$d_{I}({\bf p}_{k},\hat{\bf p})$}. Because each landmark cannot appear in all testing images, for each filter $f$, the metric only uses inference results with the top-$M$ highest activation scores on images containing the $k$-th landmark to compute {\small$D_{f,k}$}. In this way, the average of relative location deviations of all the filters in a conv-layer \emph{w.r.t.} all landmarks, \emph{i.e.} {\small${\textrm{mean}}_{f}{\textrm{mean}}_{k=1}^{K}D_{f,k}$}, measures the location instability of a CNN, where {\small$K$} denotes the number of landmarks.

\section{Network interpretability for middle-to-end learning}
\label{sec:middletoend}

Based on studies discussed in Sections~\ref{sec:explanatory} and \ref{sec:interpretableNets}, people may either disentangle representations of a pre-trained CNN or learn a new network with interpretable, disentangled representations. Such interpretable/disentangled network representations can further enable middle-to-end model learning at the semantic level without strong supervision. We briefly review two typical studies~\cite{DeepQA,holdingHands} of middle-to-end learning as follows.

\begin{figure}[t]
\centering
\includegraphics[width=\linewidth]{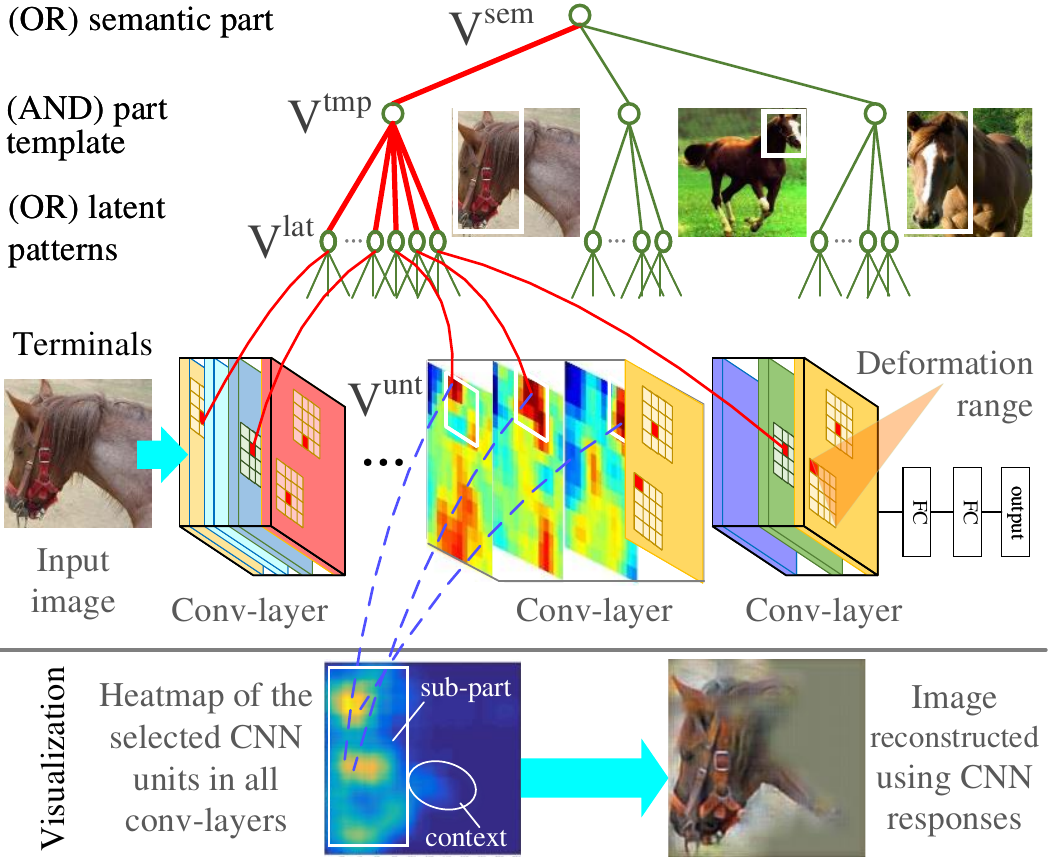}
\caption{And-Or graph grown on a pre-trained CNN as a semantic branch~{\protect\cite{DeepQA}}. The AOG associates specific CNN units with certain image regions. The red lines indicate the parse graph.}
\label{fig:AOG}
\end{figure}

\subsection{Active question-answering for learning And-Or graphs}

Based on the semantic And-Or representation proposed in \cite{CNNAoG}, \cite{DeepQA} has developed a method to use active question-answering to semanticize neural patterns in conv-layers of a pre-trained CNN and build a model for hierarchical object understanding.

As shown in Fig.~\ref{fig:AOG}, the CNN is pre-trained for object classification. The method aims to extract a four-layer interpretable And-Or graph (AOG) to explain the semantic hierarchy hidden in a CNN. The AOG encodes four-layer semantics, ranging across the \textit{semantic part} (OR node), \textit{part templates} (AND nodes), \textit{latent patterns} (OR nodes), and \textit{neural units} (terminal nodes) on feature maps. In the AOG, AND nodes represent compositional regions of a part, and OR nodes encode a list of alternative template/deformation candidates for a local part. The top part node (OR node) uses its children to represent some template candidates for the part. Each part template (AND node) in the second layer uses children latent patterns to represent its constituent regions. Each latent pattern in the third layer (OR node) naturally corresponds to a certain range of units within the feature map of a filter. The latent pattern selects a unit within this range to account for its geometric deformation.

\begin{figure}[t]
\centering
\includegraphics[width=\linewidth]{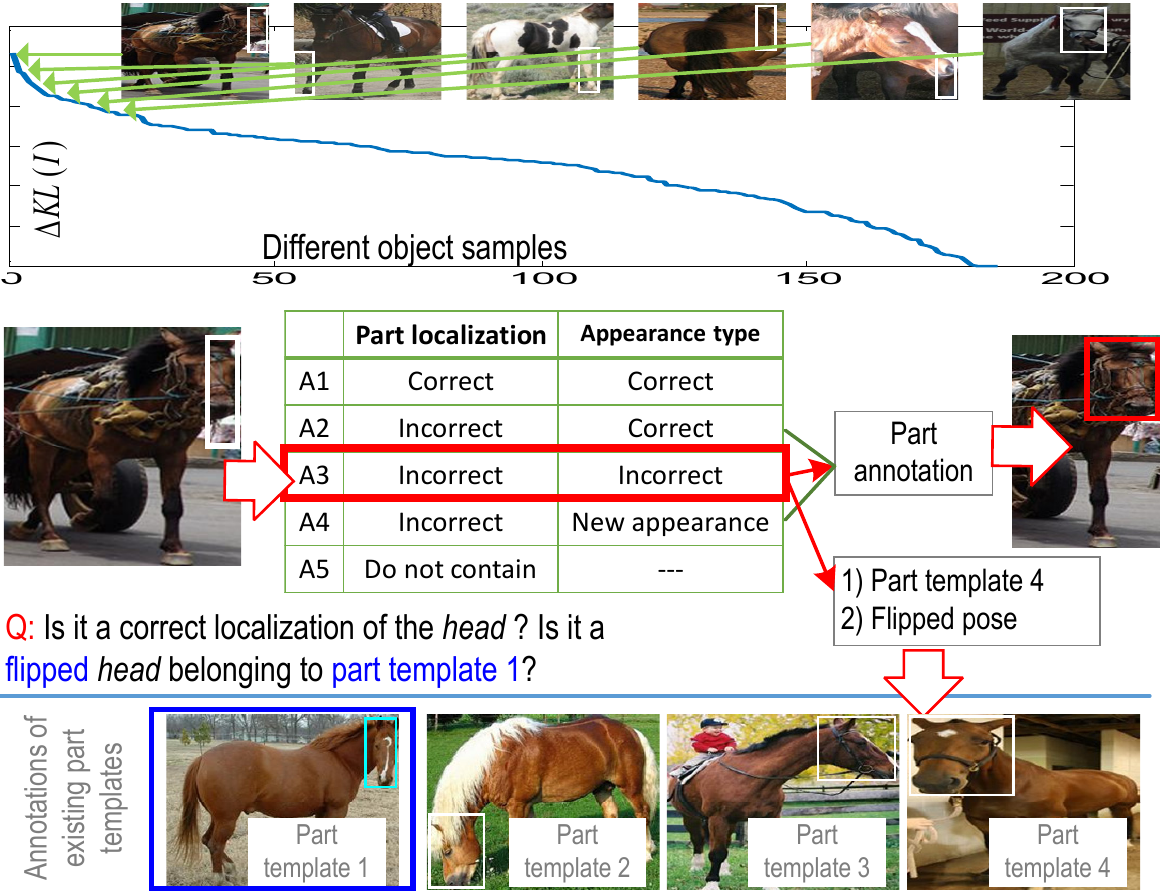}
\caption{Illustration of the QA process~{\protect\cite{DeepQA}}. (top) The method sorts and selects unexplained objects. (bottom) Questions for each target object.}
\label{fig:QA}
\includegraphics[width=\linewidth]{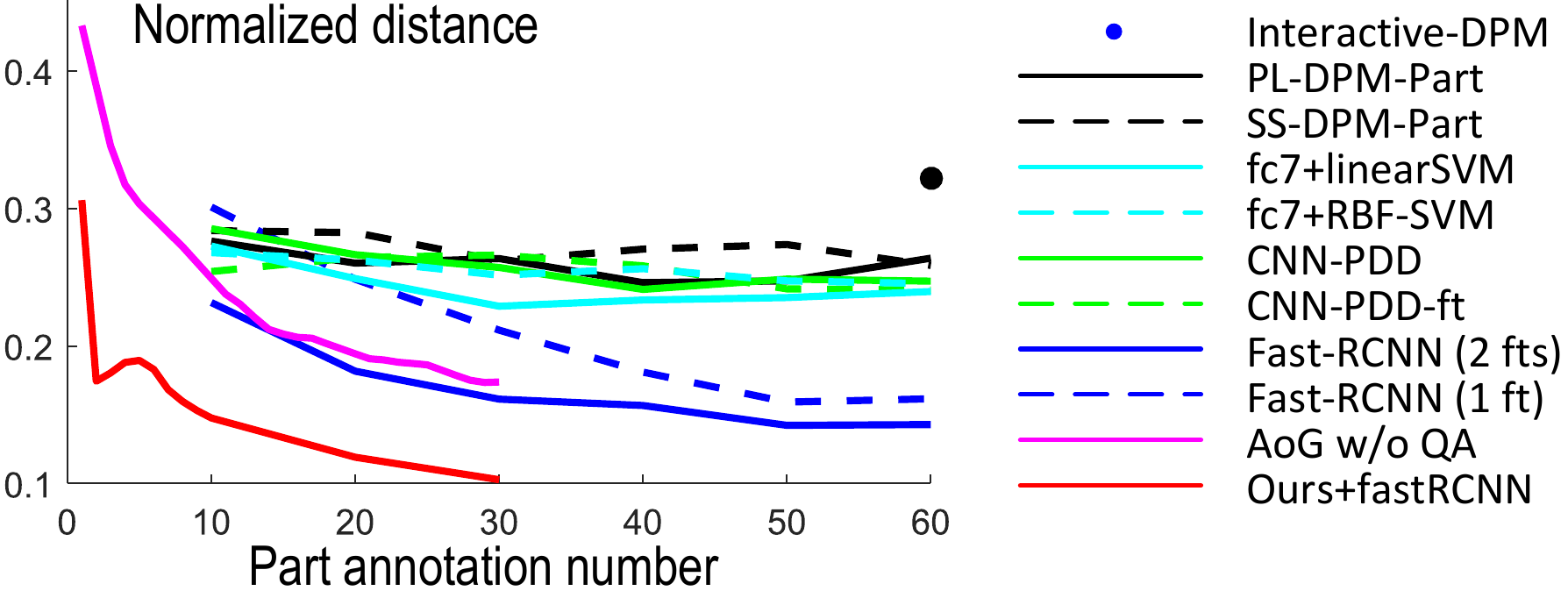}
\caption{Part localization performance on the Pascal VOC Part dataset~{\protect\cite{DeepQA}}.}
\label{fig:QA_curve}
\end{figure}

To learn an AOG, \cite{DeepQA} allows the computer to actively identify and ask about objects, whose neural patterns cannot be explained by the current AOG. As shown in Fig.~\ref{fig:QA}, in each step of the active question-answering, the current AOG is used to localize object parts among all the unannotated images. The method actively selects objects that cannot well fit the AOG, namely unexplained objects. The method predicts the potential gain of asking about each unexplained object, and thus determines the best sequence of questions (\emph{e.g.} asking about template types and bounding boxes of unexplained object parts). In this way, the method uses the answers to either refine an existing part template or mine latent patterns for new object-part templates, to grow AOG branches. Fig.~\ref{fig:QA_curve} compares the part-localization performance of different methods. The QA-based learning exhibits significantly higher efficiency than other baselines. The proposed method uses about 1/6--1/3 of the part annotations for training, but achieves similar or better part-localization performance than fast-RCNN methods.

\subsection{Interactive manipulations of CNN patterns}

\begin{figure}[t]
\centering
\includegraphics[width=\linewidth]{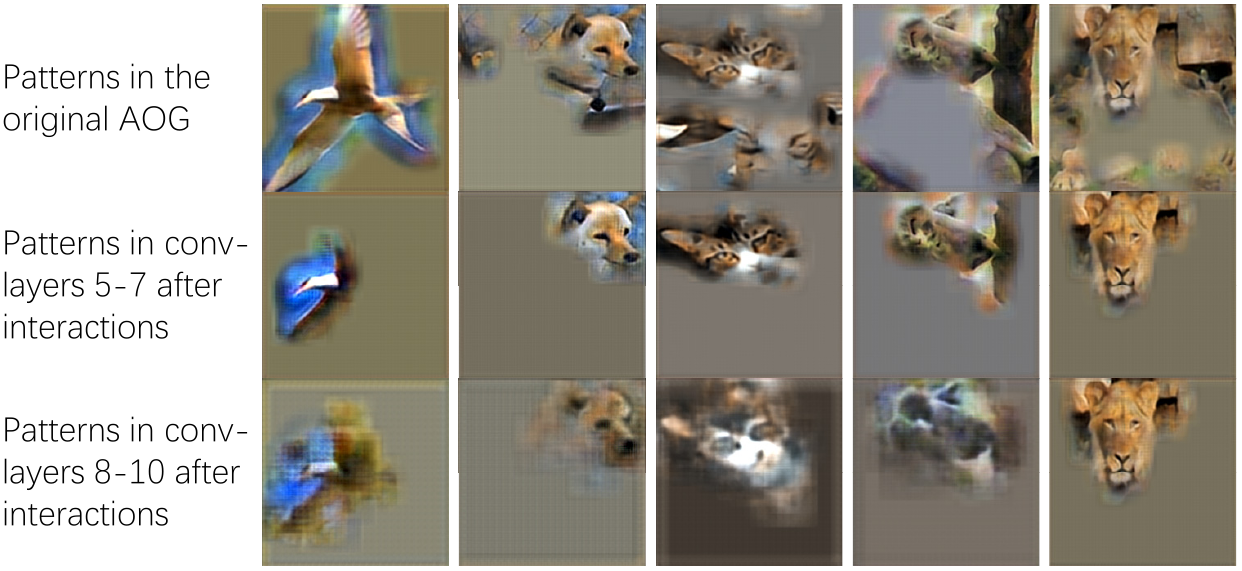}
\caption{Visualization of patterns for the head part before and after human interactions~{\protect\cite{holdingHands}}.}
\label{fig:interactiveLearn}
\end{figure}

Let a CNN be pre-trained using annotations of object bounding boxes for object classification. \cite{holdingHands} has explored an interactive method to diagnose knowledge representations of a CNN, in order to transfer CNN patterns to model object parts. Unlike traditional end-to-end learning of CNNs that requires numerous training samples, this method mines object part patterns from the CNN in the scenario of one/multi-shot learning.

More specifically, the method uses part annotations on very few (e.g. three) object images for supervision. Given a bounding-box annotation of a part, the proposed method first uses \cite{CNNAoG} to mine latent patterns, which are related to the annotated part, from conv-layers of the CNN. An AOG is used to organize all mined patterns as the representation of the target part. The method visualizes the mined latent patterns and asks people to remove latent patterns unrelated to the target part interactively. In this way, people can simply prune incorrect latent patterns from AOG branches to refine the AOG. Fig.~\ref{fig:interactiveLearn} visualizes initially mined patterns and the remaining patterns after human interaction. With the guidance of human interactions, \cite{holdingHands} has exhibited superior performance of part localization.

\section{Prospective trends and conclusions}

In this paper, we have reviewed several research directions within the scope of network interpretability. Visualization of a neural unit's patterns was the starting point of understanding network representations in the early years. Then, people gradually developed methods to analyze feature spaces of neural networks and diagnose potential representation flaws hidden inside neural networks. At present, disentangling chaotic representations of conv-layers into graphical models and/or symbolic logic has become an emerging research direction to open the black-box of neural networks. The approach for transforming a pre-trained CNN into an explanatory graph has been proposed and has exhibited significant efficiency in knowledge transfer and weakly-supervised learning.

End-to-end learning interpretable neural networks, whose intermediate layers encode comprehensible patterns, is also a prospective trend. Interpretable CNNs have been developed, where each filter in high conv-layers represents a specific object part.

Furthermore, based on interpretable representations of CNN patterns, semantic-level middle-to-end learning has been proposed to speed up the learning process. Compared to traditional end-to-end learning, middle-to-end learning allows human interactions to guide the learning process and can be applied with very few annotations for supervision.

In the future, we believe the middle-to-end learning will continuously be a fundamental research direction. In addition, based on the semantic hierarchy of an interpretable network, debugging CNN representations at the semantic level will create new visual applications.

\section*{Acknowledgement}
This work is supported by ONR MURI project N00014-16-1-2007 and DARPA XAI Award N66001-17-2-4029, and NSF IIS 1423305.

\bibliographystyle{named}
\bibliography{TheBib}

\end{document}